\documentclass{bmvc2k}

\usepackage{graphicx}
\usepackage{wrapfig}
\usepackage{amsmath}
\usepackage{amssymb}
\usepackage{booktabs}
\usepackage{adjustbox}
\usepackage{pifont}
\usepackage{makecell}
\usepackage{xcolor}
\usepackage{bm}
\usepackage{color, colortbl}
\usepackage{multirow}
\usepackage{multicol}
\usepackage{enumerate}
\usepackage{comment}

\newcommand{\xmark}{\ding{55}}
\newcommand{\gray}[1]{\color{gray}{#1}}

\definecolor{orange}{rgb}{0.98823529, 0.85098039, 0.7372549}
\definecolor{cyan}{rgb}{0.10980392, 0.45490196, 0.55294118}
\newcommand{\cc}{\cellcolor{orange}}
\renewcommand\labelenumi{(\roman{enumi})}
\renewcommand\theenumi\labelenumi

\title{Where are my Neighbors? Exploiting Patches Relations in Self-Supervised \\Vision Transformer\vspace{-5pt}}

\addauthor{Guglielmo Camporese}{guglielmo.camporese@phd.unipd.it}{1}
\addauthor{Elena Izzo}{elena.izzo@phd.unipd.it}{1}
\addauthor{Lamberto Ballan}{lamberto.ballan@unipd.it}{1}

\addinstitution{
 Visual Intelligence and Machine Perception (VIMP) Group\\
 University of Padova\\
 Padova, Italy
}

\runninghead{Camporese, Izzo, Ballan}{Exploiting Patches Relations in SSL ViT}

\begin{document}

\maketitle

\begin{abstract}
Vision Transformers (ViTs) enabled the use of the transformer architecture on vision tasks showing impressive performances when trained on big datasets. However, on relatively small datasets, ViTs are less accurate given their lack of inductive bias. To this end, we propose a simple but still effective Self-Supervised Learning (SSL) strategy to train ViTs, that without any external annotation or external data, can significantly improve the results. Specifically, we define a set of SSL tasks based on relations of image patches that the model has to solve before or jointly the supervised task. Differently from ViT, our RelViT model optimizes all the output tokens of the transformer encoder that are related to the image patches, thus exploiting more training signals at each training step. We investigated our methods on several image benchmarks finding that RelViT improves the SSL state-of-the-art methods by a large margin, especially on small datasets.
Code is available at: \url{https://github.com/guglielmocamporese/relvit}.
\end{abstract}

\section{Introduction}
\label{sec:intro}
Vision Transformer (ViT)~\cite{Dosovitskiy2021AnII} is a model that has been recently developed to address computer vision tasks, such as image classification and object detection.
It builds on the Transformer architecture~\cite{vaswani2017attention}, the state of the art in natural language processing, considering patches as parts of the image such as words are parts of a sentence.
Although ViT is a valid convolution neural networks (CNNs) competitor, showing comparable and even better results \cite{Raghu2021}, a high-level performance is obtained only when training the model on huge amounts of data.
To deal with this issue, aiming at better generalization levels, some recent works modified the attention backbone of ViT introducing hierarchical feature representation~\cite{liu2021swin}, progressively tokenization of the image~\cite{yuan2021tokens} or a shrinking pyramid backbone~\cite{wang2021pyramid}. 
Other works, inspired by BERT~\cite{Devlin2019BERTPO}, used a Self-Supervised Learning (SSL) paradigm by firstly pre-training ViT on a massive amount of unlabelled data and, subsequently, training a linear classifier over the frozen feature of the model or fine-tuning the pre-trained model to a downstream task~\cite{Caron2021EmergingPI, Chen2021ExploringSS}. 
These variants outperform similar-size ResNet trained on ImageNet, although experiments carried out on smaller datasets are still limited.

In this paper, we investigate ViTs on small datasets by introducing SSL tasks that face the problem of learning spatial relations among pairs of patches.
In particular:
\emph{(i)} We propose and investigate various SSL tasks based on image patches, naturally used by ViTs, showing the improvement optimizing all input tokens, not only the classification one;
\emph{(ii)} We show that our SSL tasks are beneficial for the model under two different settings: when the model is pre-trained from scratch on our SSL tasks and subsequently fine-tuned on classification, and when the model is jointly trained from scratch for classification and SSL tasks;
\emph{(iii)} We show that on small datasets, both ViT and its variants jointly trained from scratch on our SSL tasks reach better performance with respect to similar state-of-the-art models, and they overcome the same architectures trained with full supervision.

\begin{figure}[t]
    \centering
    \includegraphics[width=0.7\linewidth]{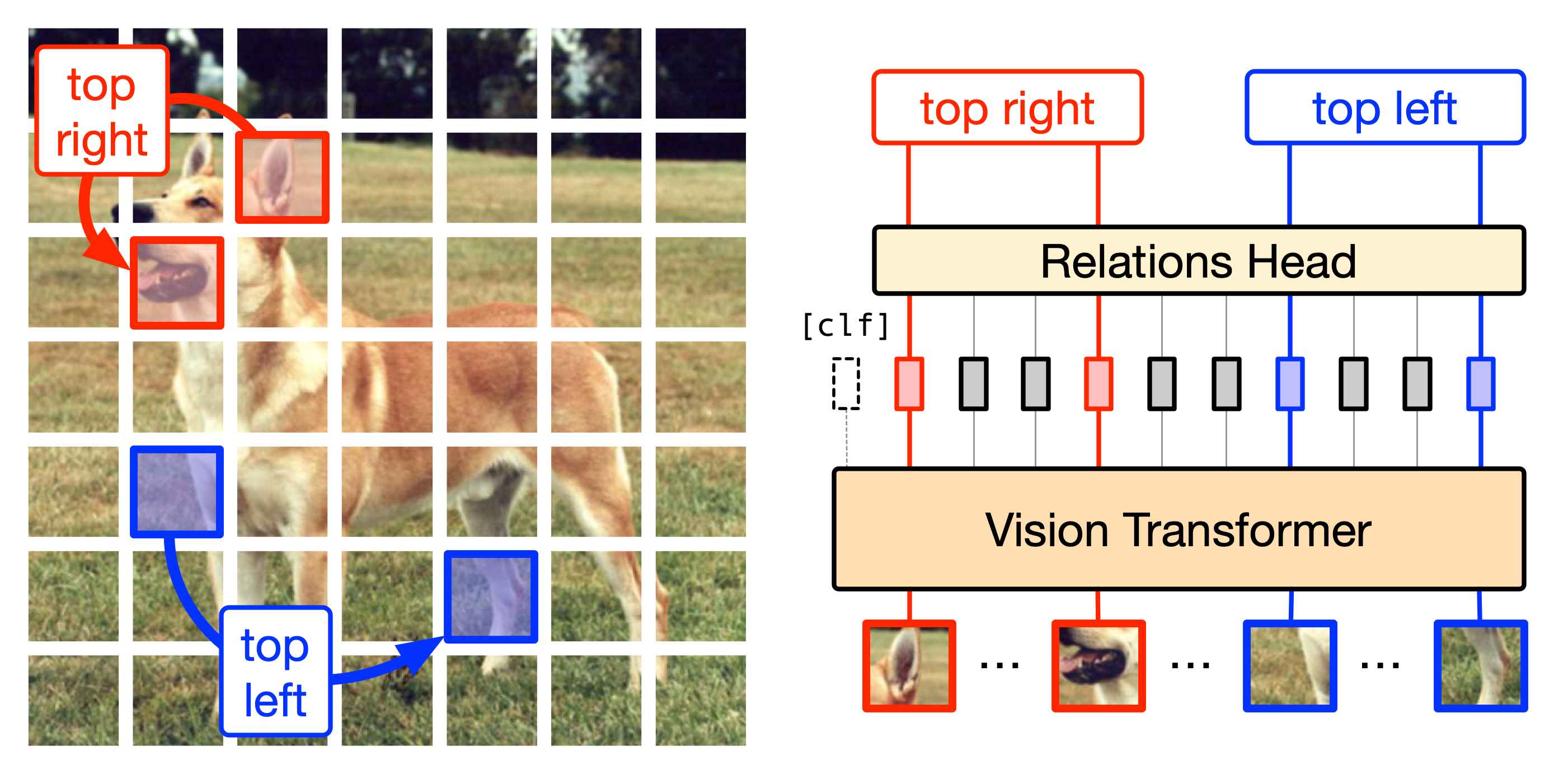} \\
    \caption{ViT splits the image into patches, and only the output classifier token that aggregates the global information of the image is directly trained with the label annotation. On the other hand, all the other tokens related to the input patches of the image are not used. Our work proposes several self-supervised tasks based on image patches used by ViT that can be beneficial for building strong internal features of the model. An example of such a task is the recognition of spatial relations among all the patches (in the picture above, only two relations examples are reported), whereas other tasks are described in Section~\ref{sec:our_method}.\label{fig:main_fig}}
\end{figure}

\section{Related Work}
\label{sec:related_work}
Our work focuses on the recently proposed Vision Transformer~\cite{Dosovitskiy2021AnII}, and it introduces several self-supervised strategies to learn internal representations for downstream tasks.
Here we highlight some of the most relevant recent efforts to move beyond fully supervised learning of vision models by exploiting self-supervised learning and pre-training strategies.
\medskip

\noindent
\textbf{Self-Supervised Learning in Vision.}
Semi-supervised learning methods for image classification share a common workflow that foresees the definition of a pretext task on unlabelled images and the subsequent evaluation of the learned features through the training of a linear classifier over the frozen representations or by the fine-tuning of the pre-trained model to a downstream task.
Previous works on Self-Supervised Learning (SSL) investigate discriminative strategies, for instance classification~\cite{Chen2020ASF, NIPS2014_07563a3f, He2020MomentumCF, Chen2020ImprovedBW, Wu2018UnsupervisedFL}, where each image has its own different class and the model is trained for discriminating them up to data perturbations. However, such modeling frameworks do not scale with the increasing number of input samples as the model is trained to discriminate between all the images. 
Other works on self-supervised learning focus on recognizing known perturbation functions on augmented images~\cite{NIPS2014_07563a3f, DBLP:conf/iclr/GidarisSK18}, by using image patches for predicting their relative positions~\cite{Doersch2015UnsupervisedVR, Noroozi2016UnsupervisedLO, liu2021efficient}, counting features~\cite{Noroozi2017RepresentationLB}, image colorization~\cite{Zhang2016ColorfulIC}, denoising autoencoders, and context encoders~\cite{Vincent_etal, Pathak2016ContextEF, Zhang2017SplitBrainAU}. 
Other recent ideas focus on contrastive methods~\cite{1640964} where the model is trained to attract positive sample pairs and at the same time to repulse negative sample pairs~\cite{Wu2018UnsupervisedFL, Ye2019UnsupervisedEL, Oord2018RepresentationLW, hjelm2018learning, henaff2020data}. 
More recent methods explore siamese architectures~\cite{Chen2020ASF, Caron2021EmergingPI, Caron2020UnsupervisedLO, Chen2021ExploringSS} where the representation of the student branch is compared and pushed towards the representation of the teacher that sees a different view of the input image.
\newline
\newline
\noindent
\textbf{Vision Transformers.}
Transformers~\cite{vaswani2017attention} are architectures designed for processing sequential data and, through the attention layers, they enable global dependencies among the input tokens. Originally, the transformer architecture shined on NLP-related tasks raising the hypothesis that even on vision could have potentially been beneficial. Despite the substantial difference between text and images, a recent breakthrough model developed for image classification, dubbed Vision Transformer~\cite{Dosovitskiy2021AnII}, adopted the transformers for images by treating the image as a sequence of patches like words are considered for phrases. This work highlights the high potential of ViTs when they are trained on large datasets. However, the high performances are not maintained when dealing with small datasets due to the lack of inductive bias of the architecture. As result, in the last few months many ViT variants emerged adapting the transformer architecture for the computer vision tasks. 
Some works replaced the rectangular backbone characterizing standard ViT with a pyramidal structure~\cite{liu2021swin, wang2021pyramid, yuan2021tokens, heo2021rethinking, wu2021cvt, fan2021multiscale} obtaining an improvement in accuracy at small/medium scale but increasing complexity and hyperparameters. Other works introduced effective design principals of CNNs into ViT~\cite{wu2021cvt, xiao2021early, chen2021visformer} in order to improve performance and robustness. 


\section{Our Method}
\label{sec:our_method}

The ViT model takes an image $\bm{x} \in \mathbb{R}^{C \times H \times W}$ as input, divides it into $N$ patches $x_p \in \mathbb{R}^{N \times (P^2 \cdot C)}$ of $C$ channels and size $(P, P)$ and makes the output prediction as follows:
\begin{equation}
    \bm{z}_0 = \big[ \bm{x}_{\text{class}}, \bm{x}_p^1\bm{E}, \dots, \bm{x}_p^N \bm{E} \big] + \bm{E}_{\text{pos}}
\end{equation}
\begin{equation}
    \bm{z}_\ell^\prime = \text{MSA}(\text{LN}(\bm{z}_{\ell-1})) + \bm{z}_{\ell-1}, \ \ \ \ell = 1 \dots L
\end{equation}
\begin{equation}
    \bm{z}_\ell = \text{MSA}(\text{LN}(\bm{z}_{\ell}^\prime)) + \bm{z}_{\ell}^\prime, \ \ \ \ell = 1 \dots L
\end{equation}
\begin{equation}
    \bm{y} = \text{LN}(\bm{z}_L^0)
\end{equation}
where $\bm{z}_\ell^i$ is the $i$-th token of the $\ell$-th layer, $\bm{E}$ is the projection matrix of the input patches, $\bm{E}_{\text{pos}} \in \mathbb{R}^{(N + 1) \times D}$ is the learned positional embedding matrix, $\text{MSA}(\cdot)$ is the multi-head self-attention layer~\cite{vaswani2017attention}, and $\text{LN}(\cdot)$ is the layer normalization block~\cite{Ba2016LayerN}. As done in BERT~\cite{Devlin2019BERTPO}, the model uses a randomly initialized classifier token $\bm{x}_{\text{class}}$ that, at the end, is taken as the reference for making the prediction $\bm{y}$.

\subsection{Learning From All Tokens}
\label{subsec:learnFromTokens}

We believe that during the learning process, extending the optimization of the classification token to all the output tokens of ViT could make the model more accurate and the training more efficient.
To this end, inspired by~\cite{Doersch2015UnsupervisedVR, Noroozi2016UnsupervisedLO, liu2021efficient}, we define a list of tasks over the output tokens $\bm{z}^i_L$ with $i=1 \dots N$ of the model: {\em spatial relations} ({\bf SpRel}), to face the problem of recognizing the spatial relation class of pairs of image patches, \textit{distances} (\textbf{Dist}) and \textit{angles} (\textbf{Ang}),  to learn a measure of distance and angle, respectively, between the input patches locations in the original image, and, finally, \textit{absolute positions} (\textbf{AbsPos}) for recognizing the 2-d location of the input image patch. More formally, referring to $\bm{z}_i$ with $i=1 \dots N$ as the $i$-th token of the $L$-th layer, given a pair of tokens $(\bm{z}_i, \bm{z}_j)$ where $\bm{z}_i$ is in position $\bm{p}_i = (x_i, y_i)$ and $\bm{z}_j$ is in position $\bm{p}_j = (x_j, y_j)$, we let the model to learn the spatial relation $r_{ij}$ defined as:

\begin{equation}
  \begin{split}
    r_{ij} = (s_x, s_y), \quad \text{where} \quad\quad
  \end{split}
  \begin{split}
    s_x \in \{ \text{L}, \text{C}, \text{R} \} \\
    s_y \in \{\text{T}, \text{C}, \text{B} \}
  \end{split}
\end{equation}

\noindent
where the partial spatial relations $\{\text{L}$, $\text{C}$, $\text{R}$, $\text{T}$, $\text{B}\}$ stand for $\{$\lq\lq left", \lq\lq center", \lq\lq right", \lq\lq top", \lq\lq bottom"$\}$. In practice, we let the model to solve the relations classification task over the set of possible spatial relations $r \in \mathcal{R}$ with $|\mathcal{R}| = 9$. \newline We define the distance between the patch locations $d_{ij}$ and the relative angle $\alpha_{ij}$ as:

\begin{minipage}[b]{0.4\textwidth}
\begin{equation}
    d_{ij} = \| \bm{p}_i - \bm{p}_j \|
\end{equation}
\end{minipage} \hfill
\begin{minipage}[b]{0.5\textwidth}
\begin{equation}
\label{eq:alpha}
    \alpha_{ij} = \cos^{-1} \left( \frac{\bm{p}_i \cdot \bm{p}_j}{\|\bm{p}_i\| \|\bm{p}_j\| + \epsilon} \right)
\end{equation}
\end{minipage}

To avoid numerical issues, in Eq.~\ref{eq:alpha} we add a small positive constant $\epsilon$ in the denominator and shift $\bm{p}_i$ and $\bm{p}_j$ by a constant $\bm{s}$. Afterward, we normalize both $d_{ij}$ and $\alpha_{ij}$ in $[-1, 1]$ and train the model to learn the normalized values.
Finally, we define the absolute position of each token $z_i$ simply as $i$ and we let the model solve the classification task over the absolute positions $\{1, \dots, N\}$.

\subsection{The Relational Vision Transformer}
\begin{figure}
     \centering
     \includegraphics[width=0.95\linewidth]{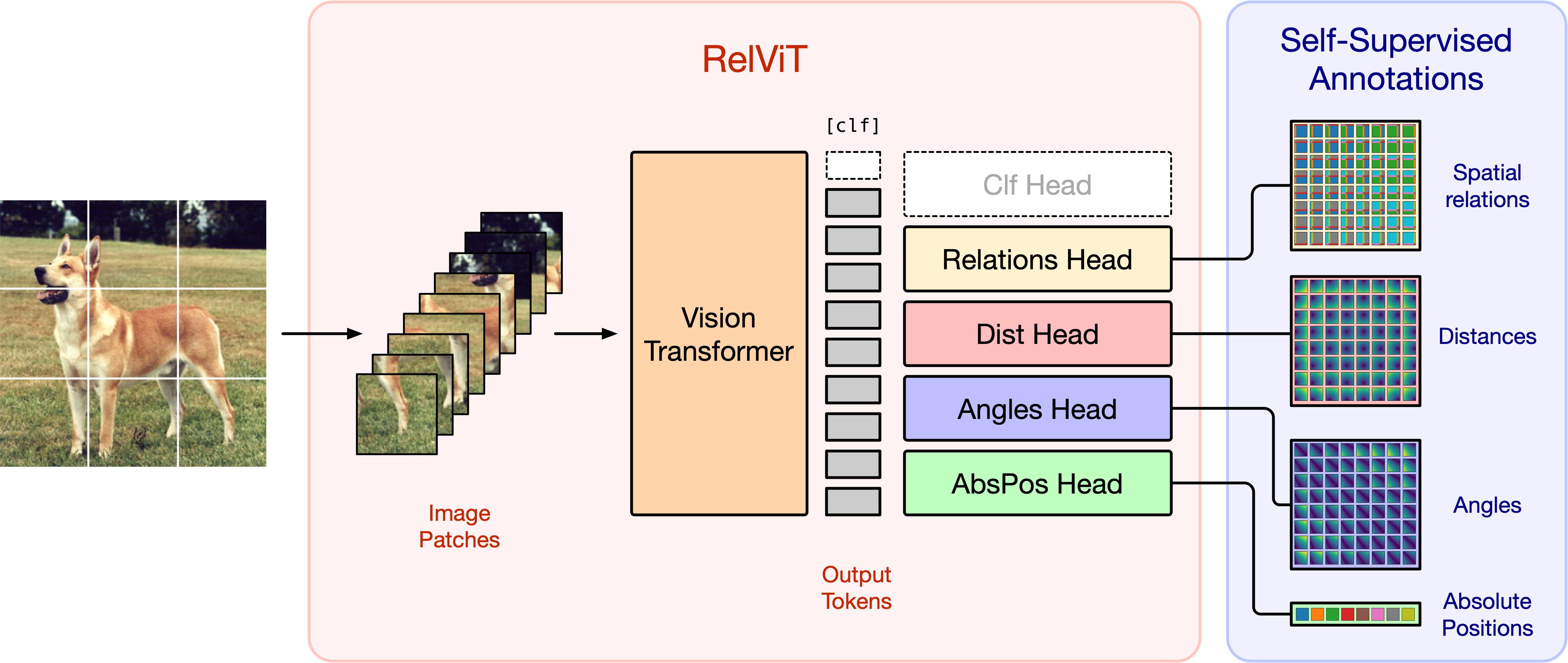}
     \caption{Our RelViT architecture. The image is partitioned into non-overlapping patches and fed to the vision transformer that produces the output tokens used by the SSL tasks.}
     \label{fig:model}
\end{figure}

Our model is depicted in Figure~\ref{fig:model}. We aim at training all the output tokens processed by the transformer backbone for all the tasks defined in~\ref{subsec:learnFromTokens}. Our architecture, dubbed RelViT, uses the standard ViT backbone equipped with specialized heads designed for solving the different tasks which take the output tokens $\bm{z} \in \mathbb{R}^{N \times D}$ of the transformer encoder as inputs. Specifically, \textit{Relations}, \textit{Dist} and \textit{Angles} heads, which process pairs of tokens, are characterized by a $\text{MSA}$ layer that produces the attention scores $\phi(\bm{z}, \bm{z}) \in \mathbb{R}^{N \times N \times K}$ described as follows:

\begin{equation}
    \phi(\bm{z}, \bm{z}) = \big[ \bm{a}_1, \dots, \bm{a}_K \big], \quad \bm{a}_i =  \frac{\bm{z}\bm{W}_{i} \bm{z}^T}{\sqrt{D}}
\end{equation}
where $\bm{a}_i \in \mathbb{R}^{N \times N}$ is the attention scores of the single attention head, $\bm{W}_i \in \mathbb{R}^{D \times D}$ is a learnable projection matrix, $K$ is the output dimension of the head, and $D$ is the token dimension.
Instead, the  \textit{AbsPos} head uses a fully connected (FC) layer that produces the logits $\phi(\bm{z}) \in \mathbb{R}^{N \times K }$ as follows:
\begin{equation}
    \phi(\bm{z}) = \bm{z} \bm{W} + \bm{b}
\end{equation}
where $\bm{W} \in \mathbb{R}^{D \times K}$ and $\bm{b} \in \mathbb{R}^K$ are respectively the weights and bias of the FC layer.\\
Given the corresponding logits, the SSL losses are computed as in the following equations:

\begin{equation}
    \mathcal{L}_{\text{sp-rel}}(\bm{z}) = \frac{1}{N^2}\sum_{i=1}^N\sum_{j=1}^N \text{CE} \big[ \phi_{\text{sp-rel}}(\bm{z}_i, \bm{z}_j), r_{ij} \big] \\
\end{equation}
\begin{equation}
    \mathcal{L}_{\text{abs-pos}}(\bm{z}) = \frac{1}{N}\sum_{i=1}^N \text{CE}\big[ \phi_{\text{abs-pos}} (\bm{z}_i), i \big]
\end{equation}
\begin{equation}
    \mathcal{L}_{\text{dist}}(\bm{z}) = \frac{1}{N^2}\sum_{i=1}^N\sum_{j=1}^N  \big\| \phi_{\text{dist}}(\bm{z}_i, \bm{z}_j) - d_{ij} \big\|^2
\end{equation}
\begin{equation}
    \mathcal{L}_{\text{angle}}(\bm{z}) = \frac{1}{N^2}\sum_{i=1}^N\sum_{j=1}^N \big\| \phi_{\text{angle}}(\bm{z}_i, \bm{z}_j) - \alpha_{ij} \big\|^2
\end{equation}
where $\text{CE}(\cdot)$ is the cross-entropy with logits loss function.
Finally, during training, we minimize the sum of all the losses of the tasks we want to solve. It is worth noticing that with this formulation, for each training step, we optimize all the combinations of pairs of patches at the same time in parallel.

\subsubsection{Mega-Patches}

Our SSL heads provide the spatial annotations thanks to a self-attention mechanism, where any output token of the ViT encoder interacts with itself and all the others in order to acquire knowledge about the spatial relations among pairs of patches.
Consequently, we believe that the increase of the number  of tokens, obtained from the non-overlapping patches on which the input image is divided, could doubly impact on the global process. First of all, there is the well-know computational effort which has a quadratic complexity with respect to the number of tokens. Secondly,  a small patch size  could reduce the improvement obtained thanks to our SSL approach since the contextual information in a patch is reduced.

\begin{wrapfigure}{r}{0.5\textwidth}
    \centering
    \includegraphics[width=\linewidth]{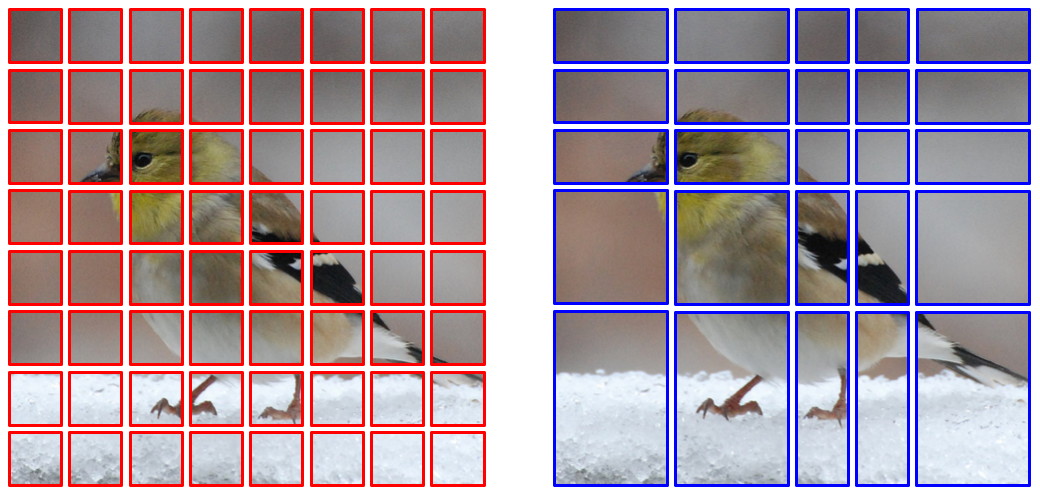} 
    \caption{Comparison between using standard patches and our proposed mega-patches. On the left, the image is divided into 64 patches, on the right it is randomly divided into 25 mega-patches using $M =5$.  \label{fig:megapatches_example}}
\end{wrapfigure}
Thus, we propose to divide the input image into $M \times M$ mega-patches which are non-overlapping rectangles whose height and width are integer multiplies of the patch size $P$. More specifically, given an image $\bm{x} \in \mathbb{R}^{C \times H \times W}$, and the number of mega-patch per side $M$, we uniformly sample $M - 1$ random indices that cut the image horizontally and vertically, producing rectangular mega-patches of different dimensions that are resized to $(P, P)$ before feeding them to our RelViT model. A qualitative difference between patches and mega-patches of an image is illustrated in Figure~\ref{fig:megapatches_example}. As shown, the mega-patches are groups of adjacent patches and we believe that their use could be beneficial for representing the image content, resulting in a better input for our RelViT model during the pre-training on the SSL tasks.

\section{Experiments}
\label{sec:experiments}
\vspace{-5pt}
We aim at demonstrating the effectiveness of learning Visual Transformers via a number of self-supervised tasks whose benefits are shown on both ViT and two of its variants: Swin~\cite{liu2021swin} and T2T-ViT~\cite{yuan2021tokens}. 

\begin{wraptable}{l}{8cm}
    \centering
    \begin{adjustbox}{width=\linewidth,center}
	\setlength{\tabcolsep}{4pt}
    \begin{tabular}{l c c c c c c} \toprule
    \textbf{Model} & \textbf{Img Res.} & \textbf{Blocks} & \textbf{Heads} & \textbf{Dim} & \textbf{\#Tokens} & \textbf{\#Params} \\
    \midrule
    ViT-S/4 & (32, 32) & 12 & 6 & 384 & 65 & 21.3M \\
    ViT-S/8 & (64, 64) & 12 & 6 & 384 & 65 & 21.4M \\
    ViT-S/32 & (224, 224) & 12 & 6 & 384 & 50 & 22.5M \\
    ViT-S/16 & (224,224) & 12 & 6 & 384 & 197 & 22.5M  \\
    ViT-S/14 & (224,224) & 12 & 6 & 384 & 257 & 22.5M \\
    \bottomrule
    \end{tabular}
    \end{adjustbox}
    \vspace{1pt}
    \caption{The used ViTs model configurations; \emph{Img Res.} is the image resolution, \emph{Blocks} and \emph{Heads} are the number of layers in the backbone and of heads in the multi-head self attention layer, \emph{Dim} is the dimension of the token in the transformer encoder, \emph{\#Tokens} is the number of patches coming from the image and \emph{\#Params} is the number of learnable parameters of the model.}
    \label{tab:models_info}
\end{wraptable}

\subsection{Datasets and Implementation Details}
\label{subsec:datasets&implDetails}
\textbf{Datasets and RelViT Configurations.}
We conducted experiments on standard image benchmarks for classification focusing on relatively small datasets providing a simple and effective way of boosting ViTs performance. In particular, we tested RelViT on CIFAR-10~\cite{Krizhevsky2009LearningML}, SVHN~\cite{Netzer2011ReadingDI}, CIFAR-100~\cite{Krizhevsky2009LearningML}, Flower-102~\cite{Nilsback2008AutomatedFC}, TinyImagenet~\cite{Le2015TinyIV} and ImageNet-100 (a subset of 100 labels from ImageNet), which have a number of train samples from 1k for Flower-102 to 130k for ImageNet-100. We used ViT-S backbone configuration~\cite{Caron2021EmergingPI} (model details in Table~\ref{tab:models_info}), while the experiments on ViT variants use Swin-T and T2T-ViT-14 backbones with the same model configurations as in~\cite{liu2021efficient}. Since our method is built upon the backbone and, once the model is trained with our SSL tasks, the additional parameters related to the heads are negligible with respect to the backbone size ($+2\%$ on the spatial relation, distance, and angle tasks and $+0.1\%$ on the absolute position task), the computational complexity added by the RelViT is negligible with respect to the visual backbone computational cost. 

\vspace{6pt}
\noindent
\textbf{Training Details.}
For all the experiments, we use the same configurations of the training parameters, except if differently mentioned. All the experiments, obtained pre-training on our SSL tasks, fine-tuning on classification or training jointly on classification and SSL tasks, are trained for $100$ epochs. Following~\cite{Caron2021EmergingPI}, we use the $\text{AdamW}$~\cite{Loshchilov2019DecoupledWD} optimizer with a cosine learning rate scheduling, linear warm-up of $10$ epochs, and $lr=0.0005$ as the maximum value for the learning rate. We found that RelViT does not need regularization, so we avoid the use of dropout, drop-path, weight decay, and label smoothing. The batch size is set to $256$ for ViT backbone and $64$ for its variants, and we use a single GPU for performing each experiment. The data augmentation for the classification problem follows the standard practice of resized crops and random horizontal flipping. Instead, when using ViT backbone and SSL tasks, each image patch is randomly resized-cropped, and following~\cite{Chen2021ExploringSS} a color shift with color-jittering is applied as well as random gray-scale perturbation.

\subsection{Experimental Results}

We investigated our RelViT model under two different scenarios: pre-training the model from scratch for our SSL tasks and subsequently fine-tuning on classification (from now on referred as to \textit{pre-training} and \textit{fine-tuning}, respectively) and training the model from scratch jointly on classification and SSL tasks (\textit{downstream-only}). We used the spatial relations and the absolute positions tasks in both pre-training and downstream-only and we removed the positional encoding in the \textit{pre-training} to avoid the shortcut learning problem, while they are used as aid whenever we want to solve the classification task, thus during both \textit{fine-tuning} and \textit{downstream-only} (more details in paragraph~\ref{subsect:ablations}). Finally, we trained all the experiments over 5 different runs in order to prove the stability of our method.

\begin{table}[ht]
    \centering
    \begin{adjustbox}{width=0.75\linewidth,center}
	\setlength{\tabcolsep}{4pt}
    \begin{tabular}{l l c c c} \toprule
     & \textbf{Backbone} & \textbf{Supervised} & \cc \textbf{RelViT} & \textbf{\color{cyan}{Improv.}} \\
    \midrule
    \textbf{CIFAR-10} & ViT-S/4 & 86.09 $\scriptstyle{\pm 0.46}$ & \cc \textbf{90.23} $\scriptstyle{\pm 0.09}$ & \small \textbf{\color{cyan}{+4.14 $\uparrow$}} \\
    \textbf{SVHN} & ViT-S/4 & 96.01 $\scriptstyle{\pm 0.07}$ & \cc \textbf{97.14} $\scriptstyle{\pm 0.03}$ & \small \textbf{\color{cyan}{+1.13 $\uparrow$}} \\
    \textbf{CIFAR-100} & ViT-S/4 & 59.19 $\scriptstyle{\pm 0.84}$ &\cc \textbf{64.99} $\scriptstyle{\pm 0.46}$ & \small \textbf{\color{cyan}{+5.85 $\uparrow$}} \\
    \textbf{Flower-102} & ViT-S/32 & 42.08 $\scriptstyle{\pm 0.29}$ & \cc \textbf{45.78} $\scriptstyle{\pm 0.75}$ & \small \textbf{\color{cyan}{+3.70 $\uparrow$}} \\
    \textbf{TinyImagenet} & ViT-S/8 & 43.19 $\scriptstyle{\pm 0.78}$ & \cc \textbf{51.98} $\scriptstyle{\pm 0.20}$ & \small \textbf{\color{cyan}{+8.79 $\uparrow$}} \\
    \textbf{Imagenet100} & ViT-S/32 & 58.04 $\scriptstyle{\pm 0.91}$ & \cc \textbf{66.46} $\scriptstyle{\pm 0.45}$ & \small \textbf{\color{cyan}{+8.42 $\uparrow$}} \\
    \bottomrule
    \end{tabular}
    \end{adjustbox}
    \vspace{1pt}
    \caption{Comparison between our RelVit model and the supervised ViT baselines on several small datasets. RelVit is pre-trained on our SSL tasks and subsequently finetuned for classification, whereas the baselines are trained on classification.}
    \label{tab:res_up-down}
\end{table}

\begin{table}[h]
    \centering
    \begin{adjustbox}{width=0.9\linewidth,center}
	\setlength{\tabcolsep}{3pt}
    \begin{tabular}{ l l c c c c} \toprule
     \textbf{Backbone} & \textbf{Method} & \textbf{CIFAR-10} & \textbf{SVHN} & \textbf{CIFAR-100} & \textbf{Flower-102} \\

    \midrule
    \multirow{4}{*}{ViT~\cite{Dosovitskiy2021AnII}} &
    & \small \gray{ViT-S/4} & \small \gray{ViT-S/4} & \small \gray{ViT-S/4} & \small \gray{ViT-S/32} \\
    & \gray{Supervised}  & \gray{\underline{85.86} $\scriptstyle{\pm 0.37}$}  & \gray{\underline{95.94} $\scriptstyle{\pm 0.05}$} & \gray{\underline{59.51} $\scriptstyle{\pm 0.99}$}  & \gray{\underline{41.74} $\scriptstyle{\pm 0.97}$} \\
    & \cc \textbf{RelViT (ours)}   & \cc \textbf{89.27 $\scriptstyle{\pm 0.23}$} & \cc \textbf{96.41 $\scriptstyle{\pm 0.39}$} & \cc \textbf{61.92} $\scriptstyle{\pm 0.19}$ & \cc \textbf{45.78 $\scriptstyle{\pm 0.75}$}  \\
    & \small (\textbf{\color{cyan}{Improv.}})      & \small (\textbf{\color{cyan}{+3.41 $\uparrow$}}) & \small (\textbf{\color{cyan}{+0.47 $\uparrow$}}) & \small (\textbf{\color{cyan}{+2.41 $\uparrow$}})            & \small (\textbf{\color{cyan}{+4.04 $\uparrow$}})   \\
    \midrule
    \multirow{4}{*}{Swin~\cite{liu2021swin}} &
    \gray{Supervised~\cite{liu2021efficient}}  & \gray{59.47} & \gray{71.6} & \gray{53.28}  & \gray{34.51} \\
    & Swin+$\mathcal{L}_{drloc}$~\cite{liu2021efficient}  & \underline{83.89} & \underline{94.23} & \underline{66.23}  & \underline{39.37} \\
    & \cc \textbf{RelSwin (ours)}   & \cc \textbf{92.33 $\scriptstyle{\pm 0.22}$} & \cc \textbf{96.64} $\scriptstyle{\pm 0.07}$ & \cc \textbf{68.69} $\scriptstyle{\pm 0.91}$ & \cc \textbf{58.47 $\scriptstyle{\pm 0.71}$}  \\
    & \small (\textbf{\color{cyan}{Improv.}})      & \small (\textbf{\color{cyan}{+8.44 $\uparrow$}}) & \small (\textbf{\color{cyan}{+2.41 $\uparrow$}}) & \small (\textbf{\color{cyan}{+2.46 $\uparrow$}})            & \small (\textbf{\color{cyan}{+19.10 $\uparrow$}})\\
    \midrule
    \multirow{4}{*}{T2T-ViT~\cite{yuan2021tokens}} &
    \gray{Supervised~\cite{liu2021efficient}}  & \gray{84.19} & \gray{95.36} & \gray{65.16}  & \gray{31.73} \\
    & T2T-ViT+$\mathcal{L}_{drloc}$~\cite{liu2021efficient}  & \underline{87.56} & \underline{96.49} & \underline{\textbf{68.03}}  & \underline{34.35}  \\
    & \cc \textbf{RelT2T-ViT (ours)}   & \cc \textbf{90.82 $\scriptstyle{\pm 0.16}$} & \cc \textbf{96.52} $\scriptstyle{\pm 0.05}$ & \cc 66.27 $\scriptstyle{\pm 0.88}$ & \cc \textbf{50.53 $\scriptstyle{\pm 1.45}$}  \\
    & \small (\textbf{\color{cyan}{Improv.}})      & \small (\textbf{\color{cyan}{+3.26 $\uparrow$}}) & \small (\textbf{\color{cyan}{+0.03 $\uparrow$}}) & \small (\textbf{\color{cyan}{-1.76 $\downarrow$}})            & \small (\textbf{\color{cyan}{+16.18 $\uparrow$}})  \\

    \bottomrule
    \end{tabular}
    \end{adjustbox}
    \vspace{1pt}
    \caption{Comparison of our method with other works on multiple backbones. The results are obtained training the model from scratch jointly on our SSL tasks and classification.}
    \label{tab:res_all_datasets}
\end{table}

Table~\ref{tab:res_up-down} shows RelViT accuracy on classification task after the \textit{fine-tuning}. Our model is compared with the standard supervised ViT, referred as \textit{Supervised}, whose results are obtained by performing both a supervised pre-training and a fine-tuning only on classification task equipping both with PEs and using the same settings described in paragraph~\ref{subsec:datasets&implDetails}.
Our approach leads to higher accuracy than the supervised one in all datasets, gaining $+8.79\%$ on $\text{Tiny-Imagenet}$, $+5.85\%$ on $\text{CIFAR-100}$ and $+4.14\%$ on $\text{CIFAR-10}$.

The first row in Table~\ref{tab:res_all_datasets} shows RelViT results on classification under the \textit{downstream-only} scenario using ViT as backbone and standard ViT as supervised baseline. Also this configuration improves the results on all the datasets up to $+4.04\%$ on $\text{Flower102}$, $+3.41\%$ on $\text{CIFAR-10}$, and $+2.41\%$ on $\text{CIFAR-100}$.
The improvements obtained in both scenarios demonstrate the effectiveness of our self-supervised tasks for image classification and the stability of our method's accuracy suggests that RelViT is not affected by large fluctuations. 

Finally, we investigated how our SSL tasks generalize across different backbones. We chose Swin~\cite{liu2021swin} and T2T-ViT~\cite{yuan2021tokens} to compare our method with~\cite{liu2021efficient} that shares a similar setting.
As reported in Table~\ref{tab:res_all_datasets}, we outperform the supervised baselines and also results reported in~\cite{liu2021efficient} up to $+19.10\%$ and $+16.18\%$ on $\text{Flower-102}$, with Swin and T2T-ViT backbones, respectively.

\subsection{Ablations}
\label{subsect:ablations}

{\bf SSL Multi-Task Ablation.}
We defined multiple SSL tasks over the image tokens and evaluated each combination of tasks in order to highlight both the importance of each sub-problem and the relations among them. 
Table~\ref{tab:res_cifar10} displays the impact of each combination of SSL tasks reporting the performance on the SSL tasks investigated during the \textit{pre-training} (columns $2^{\text{nd}}$-$6^{\text{th}}$) and the accuracy on the classification after the \textit{fine-tuning} (last column).
It is worth noticing that solving, individually, each SSL task leads to similar fine-tuning accuracy, improving the performance w.r.t the supervised baseline of $+3.00\%$ on average. Among SSL tasks, the \textit{SpRel} one is the most performing since if we rank the combinations  by the fine-tuning accuracy of the fine-tuned model, it is present in the top-3 entries, suggesting its importance for the fine-tuning classification task. Moreover, we find that the optimal combination of tasks is obtained by learning the \textit{SpRel} and the \textit{AbsPos} of the patches.

\vspace{6pt}
\noindent
{\bf Positional Embeddings and Patches Permutations.}
We investigated the shortcut learning problem in RelViT using the standard learnable 1D positional embeddings (PEs) during the self-supervised pre-training.Moreover, to avoid trivial solutions and break the dependency between PEs and SSL labels, we investigated the shuffling of the input patches before adding the PEs, permuting the rows and columns of the label matrix with the same permutation used for shuffling the input tokens. Since we aim at focusing on the role of the PEs and the patch shuffling, for simplicity we used only the \textit{SpRel} learning on the pre-training. 

\begin{wraptable}{r}{0.6\textwidth}
    \centering
    \begin{adjustbox}{width=\linewidth,center}
	\setlength{\tabcolsep}{4pt}
    \begin{tabular}{l c c c c c} \toprule
    \textbf{Method} & \textbf{\makecell{AbsPos \\ (Acc $\uparrow$)}} & \textbf{\makecell{Ang \\ (MSE $\downarrow$)}} & \textbf{\makecell{Dist \\ (MSE $\downarrow$)}} & \textbf{\makecell{SpRel \\ (Acc $\uparrow$)}} & \textbf{\makecell{Finetuning \\ (Acc $\uparrow$)}} \\
    \midrule
    \gray{Supervised}  & \gray{\xmark}     & \gray{\xmark}     & \gray{\xmark}     & \gray{\xmark} & \gray{\underline{85.38}} \\
    RelViT      & \xmark & \xmark & \xmark & 48.53  & 89.02 \\
    RelViT      & \xmark & \xmark & 10.54  & \xmark & 89.27 \\
    RelViT      & \xmark & 13.08  & \xmark & \xmark & 88.75 \\
    RelViT      & 15.24   & \xmark & \xmark & \xmark & 89.39 \\
    RelViT      & \xmark & \xmark & 10.29  & 48.91  & 89.24 \\
    RelViT      & \xmark & 12.55  & \xmark & 48.66  & 90.12 \\
    RelViT      & 15.73   & \xmark & \xmark & 48.53  & \textbf{90.13} \\
    RelViT      & \xmark & 12.58  & 10.32  & \xmark & 89.34 \\
    RelViT      & 16.03   & \xmark & 10.49  & \xmark & 89.75 \\
    RelViT      & 15.51   & 12.82  & \xmark & \xmark &  89.90 \\
    RelViT      & \xmark  & 12.49  & 10.28  & 48.98  & 88.81 \\
    RelViT      & 15.90   & \xmark & 10.39  & 48.67  & 89.80 \\
    RelViT      & 16.03   & 12.52  & \xmark & 48.80  & 89.73 \\
    RelViT      & 16.06   & 12.65  & 10.50  & \xmark & 89.67 \\
    RelViT      & 16.07   & 12.54  & 10.39  & 48.74  & 90.06 \\
    \bottomrule
    \end{tabular}
    \end{adjustbox}
    \vspace{1pt}
    \caption{Results using our SSL tasks on CIFAR-10 with ViT-S/4 backbone. The columns from the $2^{\text{nd}}$ to the $6^{\text{th}}$ report the performance on the SSL tasks whereas the last column is the accuracy of the fine-tuned models.}
    \label{tab:res_cifar10}
\end{wraptable}

\begin{wrapfigure}{l}{0.65\textwidth}
  \centering
  \includegraphics[scale=0.37]{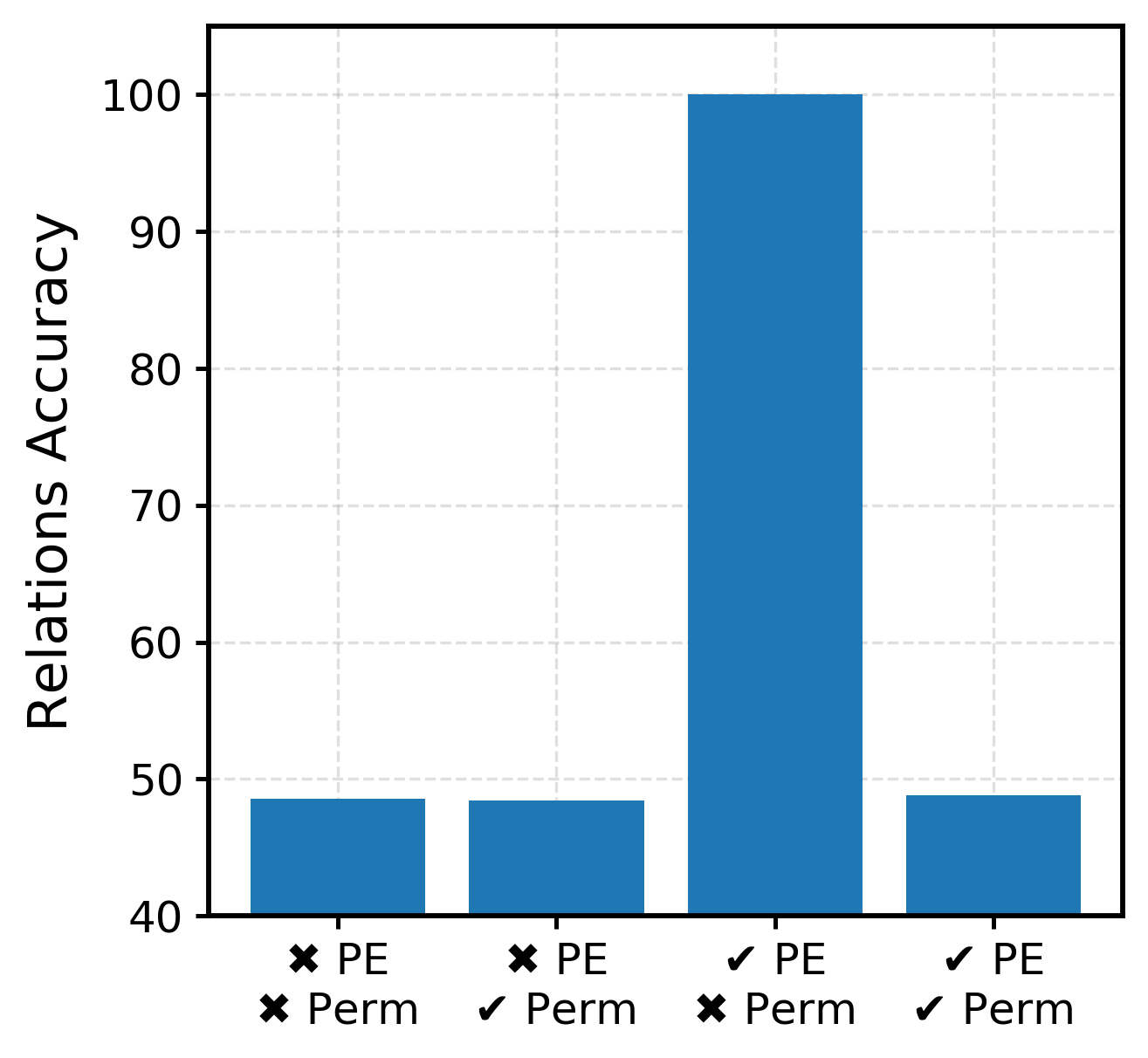}
  \includegraphics[scale=0.37]{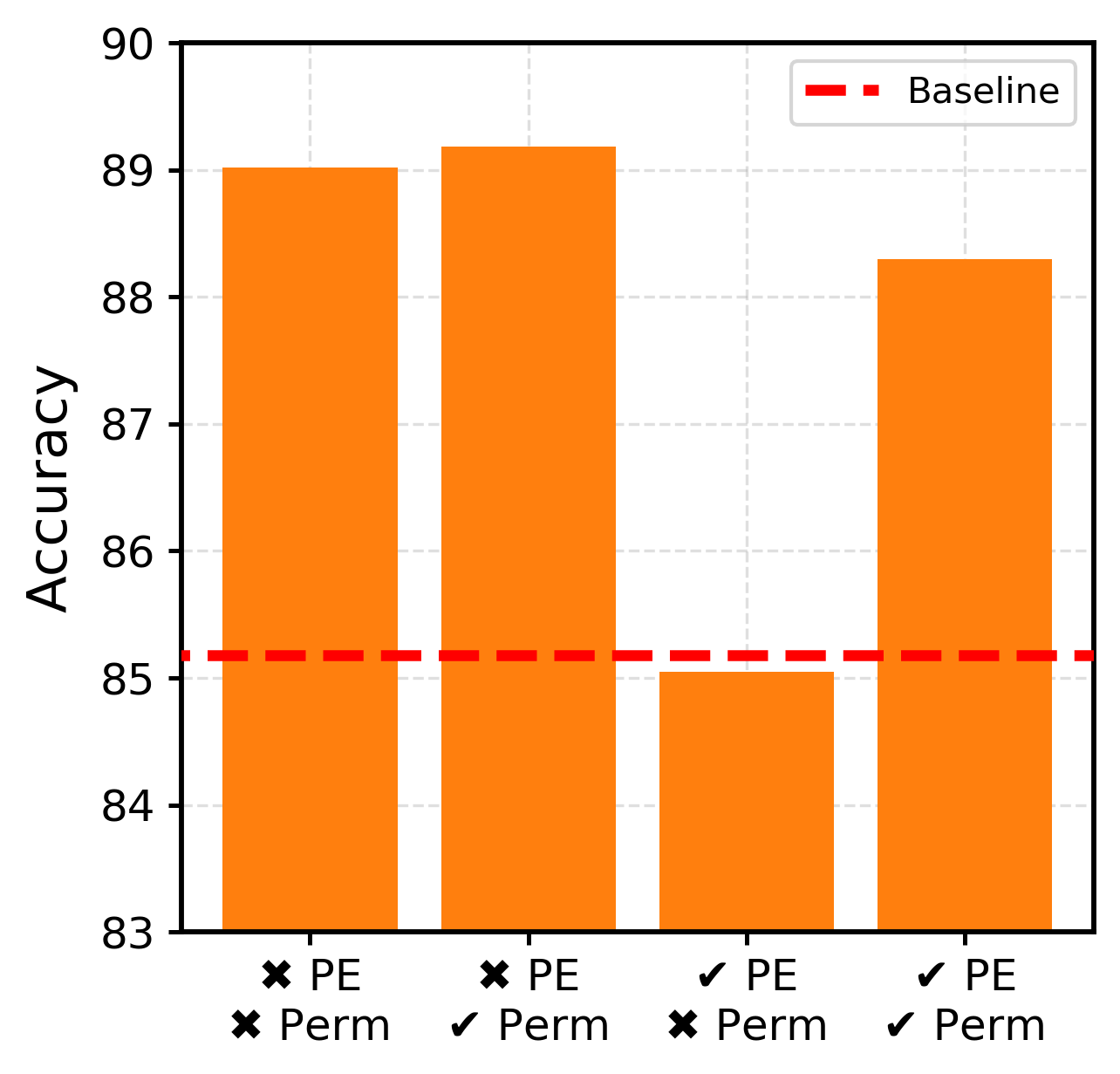}
  \caption{The role of the positional embedding (PE) and random permutations (Perm) of the input patches in RelViT during the SSL Pre-Training (left) and the Fine-Tuning (right).\label{fig:ssl_pe_perm}}
\end{wrapfigure}

As shown in Figure~\ref{fig:ssl_pe_perm}, if the model uses the PEs without shuffling during the pre-training, it solves the task with $100\%$ of \textit{SpRel} accuracy but the corresponding fine-tuned model reaches a classification accuracy comparable to a randomly initialized network (red dashed line in Fig.~\ref{fig:ssl_pe_perm}). Even though the shuffling removes the trivial solution and improves the classification accuracy, we removed PEs from RelViT during \textit{pre-training} on our SSL tasks since, in this way, we obtained the best results. Whereupon, the shuffling is optional due to the permutation equivariance property of the model. By contrast, it is important to highlight that we used PEs during both \textit{fine-tuning} and \textit{downstream-only} as well as to compute all the supervised baselines.

\vspace{6pt}
\noindent
{\bf Mega-Patches Ablation.}
We investigated our RelViT model using the mega-patches approach. To this end, the models are pre-trained from scratch using mega-patches for the spatial relations and the absolute positions tasks, and subsequently they are fine-tuned on classification using the standard patches. Table~\ref{tab:megapatches} shows RelViT accuracy after a supervised fine-tuning on ImageNet-100 and Flower-102 using three different backbones: ViT-S/32, ViT-S/16 and ViT-S/14 which differ in the patch size. \textit{Supervised} and \textit{No-Mega-Patches} are reported as references. \textit{Supervised} is obtained training only on classification, whereas \textit{No-Mega-Patches} is obtained pre-training for the spatial relations and the absolute positions without mega-patches and fine-tuning on classification. We tested several configurations by varying the parameter $M \in [3, 10]$. We observe that our RelViT model always outperforms the supervised baselines regardless of the value of the parameter $M$ underlining the effectiveness of our RelViT model. Moreover, comparing the results obtained with and without the use of the mega-patches, as expected, the mega-patches approach improves the accuracy levels as much as the patch size of the backbone is reduced suggesting that our SSL taks need more contextual information in each patch to express their potential.  

\begin{table}[h]
    \centering
    \begin{adjustbox}{width=0.8\linewidth,center}
	\setlength{\tabcolsep}{4pt}
    \begin{tabular}{ c  c c c  ccc } 
    \toprule
    & \multicolumn{3}{c}{ \cc \textbf{ImageNet-100}} & \multicolumn{3}{c}{\textbf{Flower-102}}\\
    \midrule
    $M$ & \cc ViT-S/32 & \cc ViT-S/16 & \cc ViT-S/14 & ViT-S/32 & ViT-S/16 & ViT-S/14 \\
    \midrule
    \gray{Supervised}  & \cc \gray{58.04} & \cc \gray{69.42} & \cc \gray{71.14} & \gray{41.67} & \gray{46.37} & \gray{45.39} \\
    \gray{No-Mega-Patches}  & \cc \gray{\underline{67.02}} & \cc \gray{\underline{74.74}} & \cc \gray{\underline{75.76}} & \gray{\underline{\textbf{45.49}}} & \gray{\underline{49.31}} & \gray{\underline{50.98}} \\
    10 & \cc - & \cc 75.30 & \cc 76.10 & - & 48.43 &  49.61 \\
    8 & \cc - & \cc 74.36 & \cc 76.42 &  - & 46.76 & \textbf{52.25} \\
    6 & \cc \textbf{67.26} & \cc \textbf{75.36} & \cc \textbf{76.60} & 45.39 & 49.61 & 51.67 \\
    5 & \cc 67.14 & \cc 75.26 & \cc 75.58 & 44.22 & 48.63 & 51.37 \\
    4 & \cc 66.02 & \cc 74.26 & \cc 74.24 & 44.31 & 49.31 & 50.39 \\
    3 & \cc 64.38 & \cc 73.16 & \cc 74.12 & 43.43 & \textbf{50.78} & 50.10\\
    \bottomrule
    \end{tabular}
    \end{adjustbox}
    \vspace{1pt}
    \caption{Comparison between our RelViT model using our mega-patches approach and the supervised ViT baselines. The results are obtained pre-training the model from scratch for on SSL tasks using the mega-patches and then fine-tuning on the standard patches. $M$ is the number of mega-patches per side (height or width) used during the SSL pre-training.}
    \label{tab:megapatches}
\end{table}

\section{Conclusion}
\label{sec:conclusion}
In this work, we have proposed and investigated several self-supervised tasks defined over image patches that are naturally used by ViT. Our proposed methods are straightforward to implement and effective as the RelViT shows significant improvement on all the tested datasets and in both scenarios: pre-training on SSL tasks and then fine-tuning and training jointly on SSL and SL tasks.
All the reported experiments focused on relatively small datasets, showing a simple way of making ViTs stronger even on the small dataset regime. 

\paragraph*{Acknowledgements.} 
This work was supported by the PRIN-17 PREVUE project from the Italian MUR (CUP E94I19000650001).
We also acknowledge the HPC resources of UniPD -- DM and CAPRI clusters -- and NVIDIA for their donation of GPUs used in this research.

\bibliography{references}
\end{document}